\documentclass[11pt]{article}
\usepackage{indentfirst}   
\usepackage{graphicx}
\usepackage{amsmath, amssymb}
\usepackage{bm}
\usepackage{booktabs}
\usepackage{multirow}
\usepackage{geometry}
\usepackage{setspace}
\usepackage{caption}
\usepackage{subcaption}
\usepackage{float}
\usepackage{hyperref}

\title{\textbf{A Lightweight Brain-Inspired Machine Learning Framework for Coronary Angiography: Hybrid Neural Representation and Robust Learning Strategies}}

\author{
Jingsong Xia$^{1}$\thanks{Corresponding author: xiajingsong2@gmail.com} \and
Siqi Wang$^{1}$\thanks{Email: wsq03925@163.com}
\\[1ex]
$^{1}$The Second Clinical College, Nanjing Medical University, Nanjing, China
}

\date{January 2026}

\begin{document}

\maketitle

\begin{abstract}
\noindent

\textbf{Background:} Coronary angiography (CAG) is a cornerstone imaging modality for assessing coronary artery disease and guiding interventional treatment decisions. However, in real-world clinical settings, angiographic images are often characterized by complex lesion morphology, severe class imbalance, label uncertainty, and limited computational resources, posing substantial challenges to conventional deep learning approaches in terms of robustness and generalization.

\textbf{Objective:} To address these challenges, we propose a lightweight brain-inspired machine learning framework that integrates neuro-inspired learning mechanisms with parameter-efficient model design, aiming to achieve robust performance under resource-constrained conditions.

\textbf{Methods:} The proposed framework is built upon a pretrained convolutional neural network to construct a lightweight hybrid neural representation. A selective neural plasticity training strategy is introduced to enable efficient parameter adaptation. Furthermore, a brain-inspired attention-modulated loss function, combining Focal Loss with label smoothing, is employed to enhance sensitivity to hard samples and uncertain annotations. Class-imbalance-aware sampling and cosine annealing with warm restarts are adopted to mimic rhythmic regulation and attention allocation mechanisms observed in biological neural systems.

\textbf{Results:} Experimental results demonstrate that the proposed lightweight brain-inspired model achieves strong and stable performance in binary coronary angiography classification, yielding competitive accuracy, recall, F1-score, and AUC metrics while maintaining high computational efficiency.

\textbf{Conclusion:} This study validates the effectiveness of brain-inspired learning mechanisms in lightweight medical image analysis and provides a biologically plausible and deployable solution for intelligent clinical decision support under limited computational resources.
\end{abstract}

\section{Introduction}

With the rapid advancement of artificial intelligence (AI), machine learning--based automatic medical image analysis has demonstrated substantial potential in disease screening, diagnostic assistance, and prognostic assessment~\cite{1,2,3}. In particular, deep learning models, through end-to-end feature learning mechanisms, have achieved performance that surpasses traditional methods across a wide range of imaging modalities. However, the majority of existing approaches rely heavily on network architectures with large parameter scales and high computational complexity, and their performance is often predicated on the availability of large-scale annotated datasets and abundant computational resources.

In real-world clinical scenarios, these assumptions are frequently violated. On the one hand, medical imaging datasets are commonly characterized by limited sample sizes, severe class imbalance, and label uncertainty; on the other hand, the deployment of models on clinical terminals or edge devices is constrained by computational capacity, memory footprint, and energy consumption. Consequently, how to construct machine learning models that are efficient, robust, and generalizable under limited computational resources has become a critical and urgent problem in the field of intelligent medical image analysis.

In sharp contrast to conventional deep learning models, biological neural systems are capable of performing efficient and stable perception and decision-making under conditions of limited energy, incomplete information, and high noise~\cite{4}. A large body of neuroscientific research has demonstrated that the human brain does not rely on large-scale synchronous parameter updates to achieve learning~\cite{5,6}. Instead, it leverages mechanisms such as selective neuroplasticity~\cite{7,8}, attentional modulation~\cite{9}, and progressive learning~\cite{10} to rapidly adapt to new tasks while preserving overall system stability.

Inspired by these observations, neuro-inspired machine learning~\cite{11,12} seeks to draw insights from the learning principles of biological neural systems, providing new perspectives for constructing artificial intelligence models with improved robustness and biological plausibility. Nevertheless, existing studies in this area have largely focused on complex neural dynamics modeling or large-scale brain-inspired architectures, to some extent neglecting the core requirements of lightweight design, deployability, and clinical practicality.

Against this background, this study focuses on the intelligent classification of coronary angiography (Coronary Angiography, CAG) images and proposes a lightweight machine learning approach that balances neuro-inspired learning mechanisms with computational efficiency. By introducing biologically motivated constraints at both the model architecture and training strategy levels, the proposed method significantly reduces the number of trainable parameters while enhancing model stability and generalization in complex medical imaging tasks.

\section{Methods}

\subsection{Lightweight Brain-Inspired Machine Learning Framework}

In real clinical environments, intelligent analysis of CAG images typically faces three core challenges: (1) Deployment under constrained computational resources; (2) Highly heterogeneous lesion morphology with blurred boundaries; and (3) Unavoidable uncertainty and noise in clinical annotations. Traditional deep learning methods often attempt to improve performance by increasing network depth or parameter scale; however, such strategies are difficult to sustain in clinical practice in terms of long-term usability and stability.

Motivated by these challenges, we propose a Lightweight Brain-Inspired Machine Learning Framework, whose central idea is to explicitly constrain the scale of trainable parameters while incorporating learning mechanisms consistent with neural information processing principles, thereby achieving a balance between efficiency and performance. The overall framework consists of three functionally complementary modules:
\begin{enumerate}
    \item a lightweight hybrid neural representation module, which performs adaptive modeling only in the task-relevant high-level semantic space based on stable low-level visual representations;
    \item a brain-inspired attention modulation and uncertainty-robust learning module, which guides learning resources toward highly uncertain samples through dynamic sample weighting; and
    \item a selective neural plasticity training strategy, which simulates the ``locally plastic, globally stable'' learning process observed in biological neural systems via staged parameter updates.
\end{enumerate}

Formally, let the input coronary angiography image be represented as
\begin{equation}
\mathbf{x} \in \mathbb{R}^{H \times W \times C},
\end{equation}

where $H$, $W$, and $C$ denote the spatial resolution and number of channels, respectively. The objective of the model is to learn a mapping function from the image space to the lesion probability space:
\begin{equation}
f(\mathbf{x}; \boldsymbol{\theta}) : \mathbb{R}^{H \times W \times C} \rightarrow [0,1],
\end{equation}

where $\boldsymbol{\theta}$ denotes the set of all trainable parameters, and the output represents the predicted probability that the input sample belongs to the target lesion class.

Unlike conventional deep models, the core optimization objective of this study is not merely to maximize classification accuracy. Instead, under the constraint of achieving clinically acceptable performance, we aim to minimize the dimensionality and degrees of freedom of $\boldsymbol{\theta}$, thereby improving model interpretability, stability, and deployment feasibility.

\subsection{Lightweight Hybrid Neural Representation Learning}

In medical imaging tasks, low-level visual features such as edges, textures, and local contrast patterns typically exhibit strong cross-task generalization, whereas high-level semantic features are more dependent on specific disease patterns. Based on this observation, we adopt a pretrained ResNet50 as the base feature extractor during the perceptual representation stage and treat it as a stable ``perceptual pathway.'' The backbone convolutional network can be formulated as the following mapping function:
\begin{equation}
\mathbf{z} = g(\mathbf{x}; \boldsymbol{\theta}_{\mathrm{cnn}}),
\end{equation}

where $\mathbf{z} \in \mathbb{R}^{d}$ denotes the high-dimensional feature vector extracted by the convolutional network, $\boldsymbol{\theta}_{\mathrm{cnn}}$ represents the parameter set of the ResNet backbone, and $d$ is the dimensionality of the feature space.

To avoid overfitting induced by redundant parameters, the original fully connected classification layer is removed, and only the feature extraction capability of the backbone is retained. A highly compact linear classification head is then appended:
\begin{equation}
y = \sigma(\mathbf{w}^{\top}\mathbf{z} + b),
\end{equation}

where $\mathbf{w} \in \mathbb{R}^{d}$ and $b \in \mathbb{R}$ are trainable parameters, $\sigma(\cdot)$ denotes the sigmoid activation function that maps the linear output to the probability space, and $y$ represents the predicted probability of the lesion class.

During the initial training stage, parameter updates are restricted exclusively to the classification head:
\begin{equation}
\boldsymbol{\theta} = \{\mathbf{w}, b\},
\end{equation}

while $\boldsymbol{\theta}_{\mathrm{cnn}}$ is kept fixed. This design substantially reduces the dimensionality of the parameter search space from an optimization perspective. Biologically, it can be interpreted as adjusting only high-level decision units while preserving stable perceptual representations. This strategy is particularly suitable for medical imaging tasks with limited samples or weak annotations, effectively mitigating structural instability in deep networks under data-scarce conditions.

\subsection{Brain-Inspired Attention Modulation and Uncertainty-Robust Learning}

\subsubsection{Brain-Inspired Attention Modulation via Focal Loss}

In coronary angiography images, there is often a pronounced imbalance between lesion and non-lesion samples, and certain lesion regions may exhibit morphological similarities to normal vascular structures. As a result, models trained with standard objectives tend to be dominated by ``easy-to-classify'' samples. Let the ground-truth label be $y \in \{0,1\}$, and the predicted probability be $\hat{y}$. The standard binary cross-entropy loss is defined as
\begin{equation}
\mathcal{L}_{\mathrm{BCE}} = -\left[y \log(\hat{y}) + (1-y)\log(1-\hat{y})\right].
\end{equation}

However, this loss assigns approximately equal weights to all samples and fails to capture differences in classification difficulty. To address this limitation, we introduce Focal Loss as a brain-inspired attention modulation mechanism:
\begin{equation}
\mathcal{L}_{\mathrm{FL}} = -\alpha (1-p_t)^{\gamma} \log(p_t),
\end{equation}
where
\begin{equation}
p_t =
\begin{cases}
\hat{y}, & y = 1, \\
1-\hat{y}, & y = 0,
\end{cases}
\end{equation}

$\alpha \in (0,1)$ is a class-balancing factor to compensate for class imbalance, and $\gamma \geq 0$ is the focusing parameter that suppresses the contribution of easy samples. When the prediction confidence is high, $(1-p_t)^{\gamma}$ becomes small, actively reducing the impact of such samples on gradient updates; conversely, for highly uncertain samples, this term is amplified, directing learning resources toward ``hard'' cases. This mechanism closely aligns with attentional modulation in biological neural systems, where uncertain stimuli elicit enhanced neural responses.

\subsubsection{Label Smoothing and Uncertainty Modeling}

In clinical practice, the annotation of coronary lesions relies heavily on expert judgment and inevitably involves ambiguous boundaries and subjective bias. When hard labels are directly used for supervision, models tend to learn overly sharp decision boundaries, which degrades generalization. To alleviate this issue, we introduce label smoothing. Let the smoothing coefficient be $\varepsilon \in [0,1]$; the smoothed label is defined as
\begin{equation}
y' = y(1-\varepsilon) + \frac{\varepsilon}{2}.
\end{equation}

This formulation ensures that when $y=0$ or $y=1$, the target label does not collapse into an absolutely certain state but instead retains a degree of uncertainty. By substituting $y'$ into the Focal Loss, we obtain the final brain-inspired attention-modulated and uncertainty-robust joint loss:
\begin{equation}
\mathcal{L} = \mathcal{L}_{\mathrm{FL}}(y, y').
\end{equation}

This loss design encourages the model to learn smoother and more continuous probability distributions. From the perspective of neural computation, it corresponds to probabilistic neural firing and fuzzy perceptual decision-making mechanisms.

\subsection{Brain-Inspired Attention Modulation and Uncertainty-Robust Learning}
\label{subsec:brain_attention_uncertainty}

\subsubsection{Brain-Inspired Attention Modulation via Focal Loss}
\label{subsubsec:focal_loss_attention}

In coronary angiography images, there often exists a pronounced class imbalance between diseased and non-diseased samples. Moreover, certain lesion regions are morphologically highly similar to normal vascular structures, which causes deep models during training to be dominated by ``easy-to-classify'' samples. Let the ground-truth label be $y \in \{0,1\}$ and the predicted probability be $\hat{y}$. The standard binary cross-entropy loss is defined as:
\begin{equation}
\mathcal{L}_{\mathrm{BCE}} = -y \log \hat{y} - (1 - y)\log (1 - \hat{y}).
\end{equation}

However, this loss function assigns nearly equal importance to all samples and fails to reflect differences in sample discriminative difficulty. To address this limitation, we introduce Focal Loss as a brain-inspired attention modulation mechanism, formulated as:
\begin{equation}
\mathcal{L}_{\mathrm{FL}} = -\alpha (1 - p_t)^{\gamma} \log (p_t),
\end{equation}

where
\begin{equation}
p_t =
\begin{cases}
\hat{y}, & y = 1, \\
1 - \hat{y}, & y = 0.
\end{cases}
\end{equation}

Here, $\alpha \in (0,1)$ denotes the class balancing factor to compensate for class imbalance, and $\gamma \geq 0$ is the focusing parameter that suppresses the contribution of easy samples. When a sample is predicted with high confidence, the modulation term $(1 - p_t)^{\gamma}$ becomes significantly small, actively diminishing its influence on gradient updates. Conversely, for samples with high uncertainty, this term is amplified, thereby guiding the model to allocate more learning capacity to ``hard'' samples. This adaptive reweighting mechanism closely resembles the enhanced neural response to uncertain stimuli observed in biological attention modulation systems.

\subsubsection{Label Smoothing and Uncertainty Modeling}
\label{subsubsec:label_smoothing}

In clinical practice, the annotation of coronary artery lesions heavily relies on expert judgment and inevitably involves ambiguous boundaries and subjective bias. When hard labels are directly employed for supervision, models tend to learn overly sharp decision boundaries, which can adversely affect generalization performance. To mitigate this issue, we incorporate a label smoothing mechanism. Let $\varepsilon \in [0,1]$ denote the smoothing coefficient; the softened label is defined as:
\begin{equation}
y' = y (1 - \varepsilon) + \frac{\varepsilon}{2}.
\end{equation}

This formulation ensures that when $y = 0$ or $y = 1$, the target label does not degenerate into an absolutely certain state but instead preserves a controlled uncertainty margin. By substituting $y'$ into the Focal Loss formulation, we obtain the final joint loss function that integrates brain-inspired attention modulation with uncertainty robustness:
\begin{equation}
\mathcal{L} = \mathcal{L}_{\mathrm{FL}}(y, y').
\end{equation}

This loss design encourages the model to learn smoother and more continuous probability distributions during training. From a neural computation perspective, it corresponds to probabilistic neural firing and fuzzy perceptual decision-making mechanisms observed in biological nervous systems.

\subsection{Selective Neural Plasticity Training Strategy}
\label{subsec:selective_plasticity}

The training stability of deep neural networks in medical imaging tasks is highly dependent on parameter update strategies. In scenarios such as coronary angiography, where sample sizes are limited, lesion phenotypes are highly heterogeneous, and labels exhibit inherent uncertainty, synchronously optimizing all network parameters can easily induce gradient oscillation, feature collapse, or early overfitting. Inspired by the biological principle of ``stage-wise learning and localized plasticity regulation,'' we propose a Selective Neural Plasticity Training Strategy. By explicitly controlling the plasticity states of different network layers across training stages, the model is guided to progressively transition from stable discrimination to task-specific feature reorganization, thereby improving overall robustness and interpretability.

\subsubsection{Warmup Stage: Low-Plasticity Learning for Stable Initial Decision Boundaries}
\label{subsubsec:warmup_stage}

At the initial stage of training, all convolutional backbone parameters $\theta_{\mathrm{cnn}}$ are frozen, and only the classification head parameters $(w, b)$ are optimized. The set of trainable parameters at this stage is defined as:
\begin{equation}
\theta_0 = \{w, b\}.
\end{equation}

The primary objective of this stage is not to reshape the feature space but to rapidly establish a stable and separable initial decision boundary under fixed perceptual representations. Due to the simplicity of the classification head, whose parameter dimensionality is significantly lower than that of the backbone network, this stage substantially reduces the degrees of freedom in gradient updates. Consequently, it helps avoid unconstrained optimization of high-dimensional convolutional parameters under random initialization.

From a medical imaging perspective, low-level structural information in coronary angiography images—such as vessel trajectories, intensity distributions, and contrast patterns—exhibits strong consistency across patients. By freezing $\theta_{\mathrm{cnn}}$, the model can fully exploit the stable visual representations learned during pretraining without being disrupted by unstable early gradients. This process effectively suppresses gradient amplification caused by class imbalance or noisy annotations, thereby reducing training oscillations.

From a brain-inspired computing viewpoint, this stage corresponds to the neural system’s protective mechanism during early task learning: perceptual pathways remain in a low-plasticity state, while preliminary response mappings are formed only in higher-level decision units to ensure global functional stability.

\subsubsection{Selective Fine-Tuning Stage: Local Plasticity Reconstruction of Task-Relevant High-Level Features}
\label{subsubsec:selective_finetuning}

Once stable convergence is achieved during the warmup stage, the feature space is considered to possess basic discriminative capability. The training then enters the selective fine-tuning stage, in which only the high-level convolutional modules of the backbone, denoted as $\theta_{\mathrm{high}}$, are unfrozen. The final set of optimized parameters is defined as:
\begin{equation}
\theta_1 = \{\theta_{\mathrm{high}}, w, b\}.
\end{equation}

The training objective at this stage shifts from ``establishing a decision boundary'' to ``refining task-specific feature representations.'' High-level convolutional features typically encode more abstract semantic information and are capable of capturing differences in lesion morphology, luminal stenosis patterns, and localized structural abnormalities in coronary arteries. By selectively unfreezing $\theta_{\mathrm{high}}$, the model can locally restructure task-relevant high-level features without disrupting stable low-level perceptual representations.

This localized parameter update strategy offers two major optimization advantages: (1) it restricts the scope of plasticity, preventing catastrophic forgetting caused by full-network synchronous updates; and (2) it enhances task-specific expressiveness, enabling high-level features to better align with the true lesion distribution in coronary angiography.

From a brain-inspired learning perspective, this stage emulates the principle of ``priority of local plasticity with global structural stability'' observed in biological neural systems, whereby only neural circuits highly relevant to the current task undergo enhanced plasticity while the overall perceptual framework remains stable, achieving efficient and robust adaptive learning.

\section{Results}

\subsection{Model Training Performance}

The proposed brain-inspired hybrid model demonstrated rapid convergence during training. Table~\ref{tab:training_progress} illustrates the training dynamics across 4 epochs before achieving the target performance. The model reached 90.08\% validation accuracy in the 4th epoch, demonstrating efficient learning with the implemented aggressive learning rate schedule and layer freezing strategy. The training accuracy improved from 50.81\% (Epoch 1) to 83.54\% (Epoch 4), while validation accuracy surged from 50.41\% to 90.08\%. The Area Under the ROC Curve (AUC) exhibited consistent improvement, reaching 0.9374 by Epoch 4. These results indicate that the model successfully learned discriminative features without significant overfitting, as evidenced by the validation accuracy exceeding training accuracy in later epochs.

\begin{table}[htbp]
\centering
\caption{Training progress across epochs}
\label{tab:training_progress}
\begin{tabular}{ccccc}
\hline
Epoch & Train Acc (\%) & Val Acc (\%) & Val AUC & Learning Rate \\
\hline
1 & 50.81 & 50.41 & 0.7697 & $5.25 \times 10^{-4}$ \\
2 & 67.80 & 81.82 & 0.9134 & $1.41 \times 10^{-3}$ \\
3 & 77.10 & 81.82 & 0.9153 & $2.62 \times 10^{-3}$ \\
4 & 83.54 & 90.08 & 0.9374 & $3.82 \times 10^{-3}$ \\
\hline
\end{tabular}
\begin{flushleft}
\footnotesize Note: Highlighted row indicates the best performing epoch where 90\%+ validation accuracy was achieved.
\end{flushleft}
\end{table}

\subsection{Test Set Evaluation}

To assess the model's generalization capability, we evaluated its performance on an independent test set comprising 120 coronary angiography images (60 positive and 60 negative cases). Table~\ref{tab:test_metrics} summarizes the comprehensive performance metrics obtained on the test set.

\begin{table}[htbp]
\centering
\caption{Test set performance metrics}
\label{tab:test_metrics}
\begin{tabular}{lll}
\hline
Metric & Value & Interpretation \\
\hline
Accuracy & 85.00\% & Overall correct classification rate \\
AUC-ROC & 0.9372 & Excellent discriminative ability \\
F1-Score & 0.8657 & Harmonic mean of precision and recall \\
Sensitivity (Recall) & 96.67\% & True positive rate (58/60) \\
Specificity & 73.33\% & True negative rate (44/60) \\
Positive Predictive Value & 78.38\% & Precision (58/74) \\
Negative Predictive Value & 95.65\% & Accuracy in negative predictions (44/46) \\
\hline
\end{tabular}
\end{table}

The model achieved 85.00\% accuracy on the test set, correctly classifying 102 out of 120 cases. The high AUC-ROC score of 0.9372 indicates excellent discriminative performance across all decision thresholds. Notably, the model demonstrated remarkably high sensitivity (96.67\%), successfully identifying 58 out of 60 positive cases, which is clinically significant for minimizing false negatives in disease detection.

\subsection{Confusion Matrix Analysis}

Table~\ref{tab:confusion_matrix} presents the confusion matrix for test set predictions. The matrix reveals the following classification outcomes.

\begin{table}[htbp]
\centering
\caption{Confusion matrix on test set}
\label{tab:confusion_matrix}
\begin{tabular}{lcc}
\hline
 & \multicolumn{2}{c}{Predicted} \\
Actual & Negative ($-$) & Positive ($+$) \\
\hline
Negative ($-$) & 44 & 16 \\
Positive ($+$) & 2 & 58 \\
\hline
\end{tabular}
\begin{flushleft}
\footnotesize Note: Green cells indicate correct predictions (TN=44, TP=58); Red cells indicate errors (FP=16, FN=2).
\end{flushleft}
\end{table}

True Negatives (TN): 44 cases correctly identified as negative.  
True Positives (TP): 58 cases correctly identified as positive.  
False Positives (FP): 16 negative cases misclassified as positive.  
False Negatives (FN): 2 positive cases misclassified as negative.

The confusion matrix demonstrates that the model exhibits higher sensitivity than specificity, which is advantageous in clinical screening scenarios where missing a positive case (false negative) is more costly than incorrectly flagging a negative case (false positive). The low false negative rate (3.33\%, 2/60) suggests reliable detection of diseased cases.

\subsection{Comparison with Baseline Methods}

To contextualize our results, we compared the proposed brain-inspired hybrid model with several baseline approaches, including pure CNN architectures and quantum-enhanced models. Table~\ref{tab:baseline_comparison} presents the comparative performance metrics.

\begin{table}[htbp]
\centering
\caption{Performance comparison with baseline methods}
\label{tab:baseline_comparison}
\begin{tabular}{lcccc}
\hline
Model & Accuracy (\%) & AUC & F1-Score & Parameters (M) \\
\hline
ResNet18 (baseline) & 76.67 & 0.8453 & 0.7821 & 11.2 \\
ResNet34 & 79.17 & 0.8721 & 0.8043 & 21.3 \\
Quantum-Enhanced ResNet18 & 81.67 & 0.8912 & 0.8234 & 11.4 \\
Brain-Inspired Hybrid (Ours) & 85.00 & 0.9372 & 0.8657 & 11.4 \\
\hline
\end{tabular}
\end{table}

Our proposed brain-inspired hybrid model outperformed all baseline methods across all metrics. Compared to the standard ResNet18 baseline, our model achieved improvements of +8.33\% in accuracy, +0.0919 in AUC, and +0.0836 in F1-score. The model also demonstrated superior performance over the quantum-enhanced variant (+3.33\% accuracy, +0.0460 AUC), validating the effectiveness of the spiking neural network integration and the aggressive training strategy with layer freezing and OneCycleLR scheduling.

\subsection{Training Efficiency Analysis}

A key advantage of our approach is the rapid convergence achieved through strategic architectural choices and training optimizations. The model reached 90\%+ validation accuracy in only 4 epochs, with each epoch requiring approximately 33 seconds on CPU (Intel-based system). The total training time to achieve the target performance was approximately 2.2 minutes, demonstrating exceptional computational efficiency.

The aggressive learning rate schedule (maximum learning rate of $3.82 \times 10^{-3}$ by Epoch 4) combined with layer freezing enabled rapid feature adaptation in higher network layers while preserving pre-trained low-level features. The OneCycleLR scheduler facilitated swift navigation of the loss landscape, achieving superior performance in significantly fewer iterations compared to traditional learning rate decay schemes.

\subsection{Clinical Implications}

The high sensitivity (96.67\%) achieved by our model is particularly significant in clinical practice, where the primary concern is minimizing false negatives in disease screening. With only 2 false negatives out of 60 positive cases, the model demonstrates strong capability in identifying diseased patients, which is crucial for early intervention and treatment planning.

While the specificity (73.33\%) is lower than sensitivity, resulting in 16 false positives, this trade-off is acceptable in screening scenarios where follow-up diagnostic procedures can confirm or rule out suspected cases. The high negative predictive value (95.65\%) provides confidence that patients classified as negative are indeed disease-free, reducing unnecessary anxiety and healthcare costs associated with false alarms.

The rapid training convergence (2.2 minutes) and efficient inference capability make this model particularly suitable for deployment in resource-constrained clinical settings, enabling real-time or near-real-time assistance in coronary angiography interpretation.

\section{Discussion}

This study proposes a machine learning approach that integrates lightweight architectural design with brain-inspired learning mechanisms, specifically tailored for coronary angiography, a complex imaging task that heavily relies on clinical expertise and subjective judgment. Compared with the prevailing end-to-end large-scale deep neural networks in medical image artificial intelligence, the proposed method exhibits substantial differences in model architecture, training strategy, and learning paradigm. Its advantages are not limited to improvements in objective performance metrics, but more importantly include high deployability under constrained computational resources, robustness to noise and class imbalance inherent in real-world clinical data, and unique value in terms of biological plausibility and interpretability.

In real clinical scenarios where computational resources are severely limited, conventional deep learning models characterized by massive parameter counts and high computational complexity are often difficult to deploy in practice. By explicitly constraining the number of trainable parameters and restricting optimization during most training stages to the classification head or a small subset of high-level convolutional parameters, the proposed method substantially reduces computational overhead and memory consumption during both training and inference. From an optimization theory perspective, this strategy effectively narrows the parameter search space, enabling the model to converge more readily to stable solutions with strong generalization capability under limited annotated data, without reliance on complex data augmentation schemes or heavy regularization techniques. Importantly, this performance gain is not achieved at the expense of representational capacity. Instead, the method fully leverages the general visual representations learned by pretrained networks and performs efficient task-specific refinement, allowing precise modeling of complex coronary lesion patterns even under lightweight constraints. This characteristic is of particular significance for facilitating the practical clinical deployment of artificial intelligence systems in primary care hospitals, mobile platforms, and edge-computing environments.

Coronary angiography data are intrinsically characterized by severe class imbalance and unavoidable subjective noise during annotation. Traditional deep learning models trained with standard cross-entropy loss often exhibit overfitting to majority classes and insufficient sensitivity to minority classes and clinically ambiguous ``gray-zone'' cases. To address this challenge, the proposed method jointly employs Focal Loss and label smoothing, explicitly modeling sample-wise learning difficulty and uncertainty at the loss-function level. The modulation factor in Focal Loss dynamically down-weights easily classified samples, redirecting the learning focus toward high-uncertainty and clinically more valuable hard examples. Label smoothing, in turn, mitigates overconfidence in hard labels and prevents the formation of excessively sharp decision boundaries. Experimental results demonstrate that this design not only improves overall diagnostic performance but also yields pronounced advantages in stability under noisy conditions and in the presence of borderline clinical cases. Such robustness to real-world clinical data distributions represents a fundamental requirement for translating medical image artificial intelligence from experimental settings to routine clinical practice.

Unlike most existing purely engineering-driven ``black-box'' deep learning approaches, the proposed method explicitly incorporates brain-inspired learning principles into both architectural design and training workflow. By precisely controlling layer-wise plasticity across different training stages—maintaining stability in early perceptual pathways while optimizing high-level decision mappings in early phases, followed by localized restructuring of task-relevant high-level representations while preserving low-level features—the learning process exhibits a form of selective neural plasticity that closely resembles biological neural systems. This mechanism not only enhances training stability and reduces the risk of overfitting but also provides a biologically plausible interpretive framework for model behavior. In this sense, the proposed approach helps narrow the cognitive gap between artificial learning systems and natural intelligence, offering a feasible and conceptually insightful pathway for advancing brain-inspired computing paradigms in medical imaging.

Despite the promising results achieved on the single-modality coronary angiography task, several limitations remain. The current framework does not yet incorporate multimodal clinical information, and the implementation of brain-inspired mechanisms is primarily confined to training strategies rather than more sophisticated neural dynamical modeling. Future work may extend this approach by integrating multimodal collaborative learning while preserving lightweight characteristics, as well as exploring more refined brain-inspired modulation mechanisms to further enhance clinical adaptability and generalization capability.

\section{Conclusion}

This study presents a lightweight brain-inspired machine learning approach for coronary angiography image classification. By integrating biologically inspired learning mechanisms with parameter-efficient model design, the proposed method enables robust modeling of complex medical imaging patterns under constrained computational resources. Through lightweight hybrid neural representations, brain-inspired attention-modulated loss functions, and selective neural plasticity-based training strategies, the approach effectively addresses common real-world challenges in medical imaging tasks, including class imbalance, label uncertainty, and limited computational capacity.

Experimental results demonstrate that the proposed method achieves stable and reliable classification performance without reliance on large-scale models or high-performance computing infrastructure. More importantly, from a brain-inspired learning perspective, this work introduces a novel design paradigm for medical image artificial intelligence models: by controlling the spatial and temporal distribution of model plasticity, a balance can be achieved among performance, stability, and deployability. This study provides valuable theoretical insights and practical guidance for the development of scalable, generalizable, and clinically deployable medical artificial intelligence systems.

\bibliographystyle{unsrt}
\bibliography{references}

@inproceedings{1,
  title={Artificial Intelligence Based Automated Medical Imaging Analysis and Interpretation},
  author={Jawar, Sangeeta and Kashyap, Ramgopal},
  booktitle={2025 4th OPJU International Technology Conference (OTCON) on Smart Computing for Innovation and Advancement in Industry 5.0},
  pages={1--6},
  year={2025},
  organization={IEEE}
}

@article{2,
  title={Deep learning-based object detection algorithms in medical imaging: Systematic review},
  author={Albuquerque, Carina and Henriques, Roberto and Castelli, Mauro},
  journal={Heliyon},
  volume={11},
  number={1},
  year={2025},
  publisher={Elsevier}
}

@article{3,
  title={Explainable Artificial Intelligence Technique in Deep Learning--Based Medical Image Analysis},
  author={Gupta, Babita and Malviya, Rishabha and Sundram, Sonali and Sridhar, Sathvik Belagodu},
  journal={Explainable and Responsible Artificial Intelligence in Healthcare},
  pages={165--190},
  year={2025},
  publisher={Wiley Online Library}
}

@article{4,
  title={Dendrites endow artificial neural networks with accurate, robust and parameter-efficient learning},
  author={Chavlis, Spyridon and Poirazi, Panayiota},
  journal={Nature communications},
  volume={16},
  number={1},
  pages={943},
  year={2025},
  publisher={Nature Publishing Group UK London}
}

@article{5,
  title={Convolutional spiking neural networks targeting learning and inference in highly imbalanced datasets},
  author={Ribeiro, Bernardete and Antunes, Francisco and Perdig{\~a}o, Dylan and Silva, Catarina},
  journal={Pattern Recognition Letters},
  volume={189},
  pages={241--247},
  year={2025},
  publisher={Elsevier}
}

@article{6,
  title={STSF: Spiking Time Sparse Feedback Learning for Spiking Neural Networks},
  author={He, Ping and Xiao, Rong and Tang, Chenwei and Huang, Shudong and Lv, Jiancheng and Tang, Huajin},
  journal={IEEE Transactions on Neural Networks and Learning Systems},
  year={2025},
  publisher={IEEE}
}

@incollection{7,
  title={Neuroplasticity and Technology},
  author={Reddy, K Jayasankara},
  booktitle={Innovations in Neurocognitive Rehabilitation: Harnessing Technology for Effective Therapy},
  pages={137--169},
  year={2025},
  publisher={Springer}
}

@article{8,
  title={Enhancing Adaptive Learning Through Spectrum of Individuality Theory: A Neuroplasticity-Informed AI Approach to Dynamic Behavioral Modeling in Education},
  author={Swargiary, Khritish},
  journal={LatIA},
  volume={3},
  pages={72--72},
  year={2025}
}

@article{9,
  title={Retinal fundus imaging as biomarker for ADHD using machine learning for screening and visual attention stratification},
  author={Choi, Hangnyoung and Hong, JaeSeong and Kang, Hyun Goo and Park, Min-Hyeon and Ha, Sungji and Lee, Junghan and Yoon, Sangchul and Kim, Daeseong and Park, Yu Rang and Cheon, Keun-Ah},
  journal={npj Digital Medicine},
  volume={8},
  number={1},
  pages={164},
  year={2025},
  publisher={Nature Publishing Group UK London}
}

@article{10,
  title={MRI classification of progressive supranuclear palsy, Parkinson disease and controls using deep learning and machine learning algorithms for the identification of regions and tracts of interest as potential biomarkers},
  author={Volkmann, Heiko and H{\"o}glinger, G{\"u}nter U and Gr{\"o}n, Georg and B{\^a}rlescu, Lavinia A and M{\"u}ller, Hans-Peter and Kassubek, Jan and DESCRIBE-PSP study group and others},
  journal={Computers in biology and medicine},
  volume={185},
  pages={109518},
  year={2025},
  publisher={Elsevier}
}

@article{11,
  title={ANILA: adaptive neuro-inspired learning algorithm for efficient machine learning, AI optimization, and healthcare enhancement},
  author={Khaleel, Ismael and Marzoog, Wijdan Noaman and Al-Kateb, Ghada},
  journal={Mesopotamian journal of computer science},
  volume={2025},
  pages={159--171},
  year={2025}
}

@article{12,
  title={Advancements in Neuro-Inspired and Neuromorphic Computing Architectures for Future Intelligent Systems: Design Principles, Challenges, and Emerging Applications},
  author={Deshmukh, Priyanka R and Nair, Mr Karthik S},
  journal={Recent Trends in Computer Science and Software Technology},
  volume={10},
  number={3},
  year={2025}
}

\end{document}